\documentclass[sigconf,nonacm,pbalance]{acmart}

\usepackage{graphicx}
\graphicspath{{figures/}}
\AtBeginDocument{%
  }

\setcopyright{acmlicensed}
\copyrightyear{2027}
\acmYear{2027}
\acmDOI{}
\acmConference[WWW '27]{The ACM Web Conference 2027}{May 10--14, 2027}{Dublin, Ireland}
\acmISBN{}

\begin{document}

\title{Lost in the Funnel: Selection-Induced Miscalibration in Ranking and Decision Systems}

\author{Aditya Somani}
\authornote{The first two authors contributed equally to this research.}
\email{adityasomani@microsoft.com}
\affiliation{%
  \institution{Microsoft Ads}
  \city{Bangalore}
  \state{Karnataka}
  \country{India}}

\author{Rahul Ragesh}
\authornotemark[1]
\email{rahulragesh@microsoft.com}
\affiliation{%
  \institution{Microsoft Ads}
  \city{Bangalore}
  \state{Karnataka}
  \country{India}}

\author{Vikas Joshi}
\email{vijoshi@microsoft.com}
\affiliation{%
  \institution{Microsoft Ads}
  \city{Bangalore}
  \state{Karnataka}
  \country{India}}

\author{Amit Singh}
\email{siamit@microsoft.com}
\affiliation{%
  \institution{Microsoft Ads}
  \city{Bangalore}
  \state{Karnataka}
  \country{India}}

\renewcommand{\shortauthors}{Somani et al.}

\begin{abstract}
Modern ranking and decision systems combine components with divergent objectives, and scores calibrated at the model boundary are used as calibrated probabilities throughout the funnel, across filtering, reranking, business-rule enforcement, and auctions. Yet calibration is typically validated only where the score is produced, so its breakdown where the score is used is routinely overlooked. We ask when calibration survives this selection, how inherited error differs from selection effects, and what can fail for a perfectly calibrated score at a component boundary. We study these questions through a boundary-wise diagnostic framework of residual identities. We target rubric-judged relevance rather than presentation-dependent metrics such as CTR, so calibration is measurable at every funnel layer of interest. We first present an industry-wide case study that applies our diagnosis to settings where calibration deteriorates at decision time, tracing the degradation to selection rather than model error. This diagnosis motivates \emph{Funnel-Aware Recalibration (FAR)}: post-hoc, selection-policy-aware recalibration at the boundary where scores drive decisions. In a live two-region production A/B test, FAR cuts decision-time expected calibration error (ECE) by 60--80\% and improves auction efficiency. We further adapt two public datasets to simulate industry settings, providing additional evidence for the mechanism. We plan to release the adapted datasets and their generation pipeline to support further research. Finally, we distill these findings into practical operating principles for calibrating multi-stage scoring systems.
\end{abstract}


\maketitle
\raggedbottom

\section{Introduction}
\emph{Calibration} means that a probability score $S$ for binary outcome $Y$ agrees with outcome frequencies: $\mathbb{E}[Y\mid S]=S$. For example, approximately $60\%$ of candidates assigned $0.6$ should have a positive outcome. Ranking alone does not ensure reliable magnitudes for bids, quality thresholds, or score blending in multi-signal decision systems.

Our production target is judged relevance, not a click: binary labels $Y$ ($1=$ non-defective) are assigned to query--ad pairs regardless of whether the ad is shown or engaged with. Calibration is therefore agreement with a specified relevance rubric, not an engagement frequency. Selection can still alter calibration: a score calibrated over all judged pairs need not remain calibrated over the subset that survives to a later stage. CTR instead predicts clicks on presented items, with examination and position affecting feedback. A missing click need not imply irrelevance~\cite{joachims2017}. Relevance judgments permit calibration measurement at multiple funnel layers, including candidates removed before impression, without requiring user engagement at the final stage or inferring counterfactual clicks.

Serving components change which candidates a score is applied to, so a score calibrated at model output need not remain calibrated at the point of decision. The operational question is where this error arises, and how to correct it in a principled way. Our contributions are:

\begin{enumerate}
\item \textbf{A practical diagnosis framework.} We answer three questions every industrial scoring pipeline faces: when does calibration survive selection, how can error inherited from the model be separated from error induced by selection, and what can fail even for a perfectly calibrated score? Conditional residual identities answer each question with directly measurable quantities and localize where in the funnel a correction should be applied.

\item \textbf{An industry-wide case study.} We apply the framework to a production ads funnel where calibration deteriorates at decision time. Expected calibration error (ECE) rises four- to fivefold, from $156\text{--}185$ basis points (bp; $1$\,bp $=0.01$ percentage points) at model output to $784\text{--}809$\,bp before the auction, with no change to the model. The diagnosis traces this degradation to a selection mechanism rather than to model error. Because the mechanism arises from the funnel structure, we expect the same failure in any multi-stage scoring system.

\item \textbf{Funnel-Aware Recalibration (FAR).} Motivated by the diagnosis, FAR fits a standard post-hoc calibrator that accounts for the selection policy and corrects the score before it reaches downstream decision and auction systems. It is a deployment methodology, not a new calibration estimator: calibrate where the score is used, not where it is produced. In a live two-region A/B test, FAR cuts decision-time ECE by $60\text{--}71\%$ ($470\text{--}575$\,bp) and improves auction efficiency.

\item \textbf{Public datasets and generation pipeline.} We adapt two existing public datasets to simulate industry selection settings, varying information access and selective retention, which gives independent evidence for the mechanism beyond our platform.
\end{enumerate}

\section{Related Work}
Post-hoc calibrators such as isotonic regression~\cite{zadrozny2002} are typically fit once at model output; FAR reuses these maps, and its contribution is where they are fit, not how. McMahan and Muralidharan~\cite{mcmahan2012selection} show that recalibrating on winning impressions can cycle, because recalibration changes the winners. We adopt their selection-invariance view but diagnose each boundary and correct before the auction, which weakens this loop without eliminating it, so FAR is refit periodically. Pre-impression relevance labels make this possible, since they measure calibration at every boundary, including candidates never shown, which click feedback cannot. Fan et al.~\cite{fan2023calibration} correct maximization bias in noisy CTR estimates using variance estimates and unlabeled candidates; FAR instead calibrates a frozen score on labeled survivors, since selection on information beyond the score biases survivors even without estimation noise. Bergemann et al.~\cite{bergemann2022calibrated} design calibrated CTR signals for auctions; we keep serving rules fixed and recalibrate relevance, so their click-value welfare guarantees do not carry over. Unlike propensity-based learning from click feedback~\cite{joachims2017}, FAR neither infers clicks on unshown ads nor reweights exposures; it refits calibration to the population that reaches the decision.

\section{Anatomy of a Calibration Breakdown}
\label{sec:gap}
Let $A\in\{0,1\}$ indicate survival through a component, $R=Y-S$, and let $B=B(S)$ denote fixed score bins. Write $p_b=\Pr(B=b)$, $p_b^A=\Pr(B=b\mid A=1)$, and $r_b$, $r_b^A$ for input and survivor mean residuals in bin $b$. For bins with positive survival probability,
\begin{equation}
\mathbb{E}[R\mid A=1,B]=\mathbb{E}[R\mid B]+\frac{\operatorname{Cov}(R,A\mid B)}{\Pr(A=1\mid B)}.
\label{eq:selection-gap}
\end{equation}

The survivor bias $\mathbb{E}[R\mid A=1]=\sum_b p_b^A r_b^A$ can be deconstructed as follows (proofs in Appendix~\ref{app:selection-gap-proof}):

\begin{equation}
\mathbb{E}[R\mid A=1]
=\underbrace{\textstyle\sum_b p_b r_b}_{\text{inherited error } E}
+\underbrace{\textstyle\sum_b(p_b^A-p_b)r_b}_{\text{composition shift } C}
+\underbrace{\textstyle\sum_b p_b^A(r_b^A-r_b)}_{\text{residual shift } W}.
\label{eq:decomposition}
\end{equation}

The terms decompose \emph{signed bias}, not ECE, and each points to a different remedy:
\begin{itemize}
\item \textbf{Inherited error $E$} is the bias already present in the input population, before the boundary acts. It reflects upstream miscalibration and can be corrected upstream.
\item \textbf{Composition shift $C$} arises because survivors occupy score bins in different proportions than the input. Selection creates no new per-bin error; it reweights existing bin errors. $C$ vanishes if the input is calibrated in every bin or if selection leaves the bin mix unchanged.
\item \textbf{Residual shift $W$} arises within bins: among candidates with similar scores, selection retains those whose outcomes differ systematically from the bin average. It equals the weighted covariance term of Eq.~\eqref{eq:selection-gap}, is created by the boundary itself, and can arise even for a perfectly calibrated input. Only a selection-policy-aware correction, fit on the survivors the policy actually delivers, can remove it.
\end{itemize}

\paragraph{Conditions for transfer.} By Eq.~\eqref{eq:selection-gap} with $S$ in place of $B$, a calibrated input remains calibrated on survivors if and only if $\operatorname{Cov}(R,A\mid S)=0$ (Appendix~\ref{app:transfer}). This holds when:
\begin{enumerate}
\item \textbf{Score-only selection:} $A$ depends only on $S$ and independent randomness (e.g., thresholds, top-$k$).
\item \textbf{Same-information selection:} $S=\mathbb{E}[Y\mid X]$ and $A$ depends only on $X$ and independent randomness.
\item \textbf{No extra outcome information:} more generally, $\mathbb{E}[Y\mid S,A]=\mathbb{E}[Y\mid S]$, implied by $A\perp Y\mid S$, paralleling the condition of McMahan and Muralidharan~\cite{mcmahan2012selection}.
\end{enumerate}

Transfer fails when the selector uses information predictive of $Y$ beyond $S$. If the input is miscalibrated, score-only selection shifts aggregate bias through $C$; with coarse bins, it can also induce $W$. Under extra information, even a calibrated upstream score cannot prevent $W$; only a selection-aware correction can remove it.

\paragraph{Applying the diagnosis.} We apply Eq.~\eqref{eq:decomposition} at each consecutive boundary with a common score definition and bins. Separate labeled samples suffice without row linkage, provided they represent the input and survivors of the same transition; otherwise, temporal or labeling changes can masquerade as selection. Since signed shifts can offset, small aggregate bias does not imply small bin-level error.

\section{Calibrate Where Scores Are Used}
\label{sec:far}
FAR fits an existing post-hoc map $g(S)$ on the survivors of a decision boundary rather than on the model's input population. Because survivors reflect the selection policy, the fit absorbs the composition shift $C$ and residual shift $W$, which upstream recalibration cannot remove, as well as the inherited error $E$ among survivors; no prior attribution is required. FAR leaves the model and serving rules unchanged, so it can be deployed as a lightweight layer on a frozen score. A practical recipe is:
\begin{enumerate}
\item \textbf{Declare the decision boundary.} Identify where probabilities drive decisions, and calibrate on the candidates that reach that point, since selection changes calibration. In our funnel this is the pre-auction boundary, where scores set quality thresholds and enter auction ranking.
\item \textbf{Fit under the selection policy.} Learn $g$ on labeled samples representative of the survivors the current policy delivers, with a consistent outcome definition, holding out separate samples for evaluation.
\item \textbf{Validate at the decision boundary.} Compare residuals and ECE on held-out survivors in $g(S)$ bins, and assess later boundaries separately using a common score and bins (Section~\ref{sec:gap}).
\end{enumerate}

A fit at one boundary calibrates the next boundary's input, driving $E$ and $C$ there toward zero, but later selection can create new $W$, so calibration must be rechecked downstream. As the correction and evolving filters, models, and policies reshape survivors, $g$ should be refit and monitored periodically.

\section{Industry Evaluation}
\label{sec:industry}
\begin{figure}[t]
  \centering
  \includegraphics[width=\linewidth]{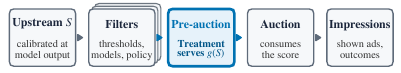}
  \caption{Serving stack with FAR $g(S)$ applied before the auction.}
  \Description{Schematic of the serving stack: the upstream relevance model produces score S, candidates pass through filtering, and the pre-auction FAR correction g(S) is applied before the auction and impression stages.}
  \label{fig:stack}
\end{figure}

Our system is a large-scale commercial sponsored ads platform. An upstream relevance model assigns $S$, evaluated against the binary relevance labels $Y$ defined above; filtering precedes auctions in which absolute scores drive quality-threshold filtering and auction dynamics (Fig.~\ref{fig:stack}).

A 20-day A/B test randomizes 5\% of search requests in two regions: Control uses $S$, Treatment uses $g(S)$ at the pre-auction boundary, all else unchanged. An isotonic map~\cite{zadrozny2002} is fit on about 500k labeled pre-auction candidates from a prior week, pooled across regions; we sample 25k candidates per region at the upstream, pre-auction, and impression boundaries. The \emph{signed gap} $\mathbb{E}[Y-S]$ (pp) shows bias direction and decomposes via Eq.~\eqref{eq:decomposition} but can cancel across bins; $\mathrm{ECE}=\sum_b p_b\,|\bar{y}_b-\bar{s}_b|$ (bp) averages absolute bin errors, bounding $|\text{gap}|$ from above. We use 50 equal-mass bins and two-sided $p$-values from 1{,}000 candidate-level bootstrap resamples.

\begin{figure}[t]
  \centering
  \includegraphics[width=\linewidth]{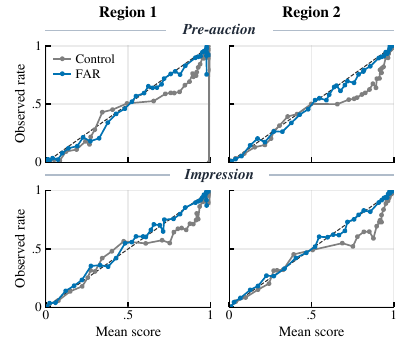}
  \caption{Reliability (50 bins): Control (gray) overstates high scores; FAR (blue) tracks the diagonal.}
  \Description{Four reliability graphs in a two-by-two grid. Columns show Regions 1 and 2; rows show pre-auction and impression populations. Each graph compares gray Control and blue FAR curves using 50 equal-mass bins. Dashed diagonals mark equal mean score and observed relevance rate.}
  \label{fig:reliability}
\end{figure}

\begin{table}[t]
  \centering
  \caption{Signed-gap decomposition at filtering in Control; residual selection shift $W$ drives most of the $-7.3/-7.8$\,pp error. \textbf{\textcolor[HTML]{0072B2}{Blue}}: $W\neq0$, bootstrap $p\le0.001$.}
  \label{tab:filtering}
  \begin{tabular}{llrrrr}
    \toprule
    Boundary & Reg. & $E$ & $C$ & $W$ & $=$ Out \\
    \midrule
    Filtering (Ctrl) & R1 & $+0.1$ & $+0.2$ & \textbf{\textcolor[HTML]{0072B2}{\boldmath$-$7.5}} & $-7.3$ \\
     & R2 & $-0.8$ & $-0.1$ & \textbf{\textcolor[HTML]{0072B2}{\boldmath$-$7.0}} & $-7.8$ \\
    \bottomrule
  \end{tabular}
\end{table}

\begin{table}[t]
  \centering
  \caption{Production A/B by region: funnel ECE (bp; upstream $S$: $156/185$) and online $\Delta\%$. FAR cuts decision-time ECE by $60\text{--}80\%$. \textbf{\textcolor[HTML]{0072B2}{Blue}}: $p<0.001$; \textcolor[HTML]{5F6B7A}{gray$^{\text{ns}}$}: $p>0.05$.}
  \label{tab:ab}
  \begin{tabular}{llrrrr}
    \toprule
    Boundary & Reg. & Ctrl & FAR & $\Delta$ & $\Delta\%$ \\
    \midrule
    Pre-auction & R1 & $784$ & $314$ & \textcolor[HTML]{0072B2}{$-470$} & \textbf{\textcolor[HTML]{0072B2}{\boldmath$-$60\%}} \\
     & R2 & $809$ & $234$ & \textcolor[HTML]{0072B2}{$-575$} & \textbf{\textcolor[HTML]{0072B2}{\boldmath$-$71\%}} \\
    Impression & R1 & $633$ & $233$ & \textcolor[HTML]{0072B2}{$-400$} & \textbf{\textcolor[HTML]{0072B2}{\boldmath$-$63\%}} \\
     & R2 & $711$ & $143$ & \textcolor[HTML]{0072B2}{$-568$} & \textbf{\textcolor[HTML]{0072B2}{\boldmath$-$80\%}} \\
    \midrule
    Online $\Delta\%$ & \multicolumn{2}{r}{Ad coverage} & CTR & Quick-back & RPM \\
    \midrule
    R1 & \multicolumn{2}{r}{\textbf{\textcolor[HTML]{0072B2}{\boldmath$-$3.1\%}}} & \textbf{\textcolor[HTML]{0072B2}{\boldmath$+$1.0\%}} & \textbf{\textcolor[HTML]{0072B2}{\boldmath$-$1.2\%}} & \textcolor[HTML]{5F6B7A}{$+$0.2\%$^{\text{ns}}$} \\
    R2 & \multicolumn{2}{r}{\textbf{\textcolor[HTML]{0072B2}{\boldmath$-$9.2\%}}} & \textbf{\textcolor[HTML]{0072B2}{\boldmath$+$2.0\%}} & \textbf{\textcolor[HTML]{0072B2}{\boldmath$-$2.4\%}} & \textcolor[HTML]{5F6B7A}{$-$0.1\%$^{\text{ns}}$} \\
    \bottomrule
  \end{tabular}
\end{table}

\paragraph{Funnel diagnosis.}
In both regions, upstream $S$ is well calibrated ($156/185$\,bp ECE), yet Control ECE jumps to $784/809$\,bp after filtering, with reliability $6\text{--}24$\,pp below the diagonal at the top of the score range (Fig.~\ref{fig:reliability}). Table~\ref{tab:filtering} shows why: inherited error and composition shift are negligible, while residual shift $W=-7.5/-7.0$\,pp drives a $-7.3/-7.8$\,pp pre-auction calibration gap. The model never changes: almost all of the four- to fivefold degradation arises in the serving stack, within score bins, where no upstream refit can reach it, motivating pre-auction FAR.

\paragraph{FAR calibration gains.}
FAR cuts pre-auction ECE by $60/71\%$ and impression ECE by $63/80\%$ (Table~\ref{tab:ab}), with reliability curves indicating significantly improved calibration (Fig.~\ref{fig:reliability}). These gains come from a single selection-policy-aware monotone map, fit on one week of pre-auction survivors, with no model retraining and no change to filtering or auction rules.

\paragraph{Auction efficiency.}
Ad coverage falls $3.1/9.2\%$, mainly by removing defective ads that Control's overstated scores let through. CTR rises $1.0/2.0\%$ and the quick-back rate, a dissatisfaction proxy, falls $1.2/2.4\%$, with no significant RPM change (Table~\ref{tab:ab}). Users therefore see fewer but more relevant ads at no measurable revenue cost: once scores are calibrated where they are used, quality thresholds act as intended.

\section{Supporting Public Replay}
\label{sec:replay}
Production funnel logs are proprietary, and public logs contain only served impressions, so the selection between calibration and decision cannot be observed. We therefore simulate it on public data, controlling the selection policy, strength, and information. We adapt KuaiRand-1K~\cite{gao2022} (6.7M impressions; $S$: Click, selector $D$: Long View) and Criteo Attribution~\cite{diemert2017} (16.5M; $S$: Click, $D$: Conversion) in three steps:
\begin{enumerate}
\item \textbf{Upstream calibration and regimes.} On a 60/20/20 train/calibration/test split, gradient-boosted trees fit $S$ and $D$; $S$ is isotonic-calibrated and frozen. $D$ sees all features, while $S$ sees all features (\emph{shared information}) or static inputs only (\emph{information gap}~\cite{xu2020}, withholding history/recency/cost to proxy later-stage signals).
\item \textbf{Selection simulation.} On the test split, 20 draws generate survivor datasets with survival probability $\sigma(\alpha+\beta(2\operatorname{rank}(D)-1))$, $\alpha$ fixing $10\%$ retention and $\pm\beta$ favoring Top-/Bot-$D$; a \emph{Hard} $10\%$ cutoff on $D$~\cite{fan2023calibration} is the $\beta\to\infty$ limit. 
\item \textbf{Survivor FAR and evaluation.} Survivors are split 50/50 to fit isotonic FAR $g(S)$ and evaluate it with the ECE and $\Delta_{\mathrm{gap}}$ of Section~\ref{sec:industry}, using user-clustered tests.
\end{enumerate}

\begin{figure}[t]
  \centering
  \caption{Replay signed-gap change from input to survivors by selection strength $\beta$ (Hard: exact cutoff). Gaps vanish at $\beta=0$ and grow with $\beta$. Bands: 95\% CIs; gray: random.}
  \includegraphics[width=\linewidth]{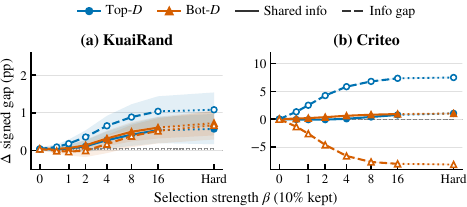}
  \Description{Two panels of survivor-minus-input signed gap in percentage points across selection strength beta at 10 percent expected retention (including hard cutoffs), with KuaiRand on the left and Criteo on the right. Blue circles favor top D ranks; orange triangles favor bottom D ranks. Solid curves use shared information and dashed curves information gaps. Shading shows pointwise user-cluster confidence intervals; gray traces show random controls. Criteo information-gap selection produces large shifts in opposite directions.}
  \label{fig:replay}
\end{figure}

\begin{table}[t]
  \centering
  \caption{Public replay $\Delta_{\mathrm{gap}}$ and ECE ($\beta=8$, 10\% kept, bp; $(r)$: random ECE). Selection shifts the gap in all runs; survivor-fit FAR removes it (post-FAR $|\Delta_{\mathrm{gap}}|\le 4$, $p>0.44$), cutting Criteo information-gap ECE by $90/93\%$. \textbf{\textcolor[HTML]{0072B2}{Blue}}: $p<0.05$; \textcolor[HTML]{5F6B7A}{gray$^{\text{ns}}$}: $p>0.05$.}
  \label{tab:replay}
  \begin{tabular}{llrrrrr}
    \toprule
    Dataset $(r)$ & Policy & $\Delta_{\mathrm{gap}}$ & Pre & FAR & $\Delta$ & $\Delta\%$ \\
    \midrule
    \multicolumn{7}{c}{\textbf{\emph{Shared information ($S, D$ share features; $(r)$: random ECE)}}} \\
    KuaiRand ($92$) & Top-$D$ & \textbf{\textcolor[HTML]{0072B2}{\boldmath$+$41}} & $90$ & $83$ & \textcolor[HTML]{5F6B7A}{$-7$} & \textbf{\textcolor[HTML]{5F6B7A}{\boldmath$-$8\%$^{\text{ns}}$}} \\
     & Bot-$D$ & \textbf{\textcolor[HTML]{0072B2}{\boldmath$+$50}} & $71$ & $53$ & \textcolor[HTML]{0072B2}{$-18$} & \textbf{\textcolor[HTML]{0072B2}{\boldmath$-$25\%}} \\
    Criteo ($59$) & Top-$D$ & \textbf{\textcolor[HTML]{0072B2}{\boldmath$+$39}} & $92$ & $72$ & \textcolor[HTML]{0072B2}{$-20$} & \textbf{\textcolor[HTML]{0072B2}{\boldmath$-$22\%}} \\
     & Bot-$D$ & \textbf{\textcolor[HTML]{0072B2}{\boldmath$+$81}} & $97$ & $54$ & \textcolor[HTML]{0072B2}{$-43$} & \textbf{\textcolor[HTML]{0072B2}{\boldmath$-$44\%}} \\
    \midrule
    \multicolumn{7}{c}{\textbf{\emph{Information gap ($D$ sees features withheld from $S$)}}} \\
    KuaiRand ($90$) & Top-$D$ & \textbf{\textcolor[HTML]{0072B2}{\boldmath$+$88}} & $121$ & $91$ & \textcolor[HTML]{0072B2}{$-30$} & \textbf{\textcolor[HTML]{0072B2}{\boldmath$-$25\%}} \\
     & Bot-$D$ & \textbf{\textcolor[HTML]{0072B2}{\boldmath$+$38}} & $64$ & $56$ & \textcolor[HTML]{5F6B7A}{$-8$} & \textbf{\textcolor[HTML]{5F6B7A}{\boldmath$-$13\%$^{\text{ns}}$}} \\
    Criteo ($61$) & Top-$D$ & \textbf{\textcolor[HTML]{0072B2}{\boldmath$+$680}} & $685$ & $70$ & \textcolor[HTML]{0072B2}{$-615$} & \textbf{\textcolor[HTML]{0072B2}{\boldmath$-$90\%}} \\
     & Bot-$D$ & \textbf{\textcolor[HTML]{0072B2}{\boldmath$-$770}} & $768$ & $53$ & \textcolor[HTML]{0072B2}{$-715$} & \textbf{\textcolor[HTML]{0072B2}{\boldmath$-$93\%}} \\
    \bottomrule
  \end{tabular}
\end{table}

\paragraph{What the replay shows.}
These settings span the transfer conditions of Section~\ref{sec:gap}. \emph{$S$-only selection} preserves calibration, as score-only selection predicts, and at $\beta=0$ (Fig.~\ref{fig:replay}), where survival ignores $D$, the signed gap vanishes. Under \emph{shared information}, small but significant gaps remain, because a fitted $S$ only approximates $\mathbb{E}[Y\mid X]$, leaving residual outcome information for the selector. Under an \emph{information gap}, Criteo gaps grow roughly tenfold, comparable to the $-7.5/-7.0$\,pp within-bin shift at production filtering.

Since $\Delta_{\mathrm{gap}}$ and survivor ECE depend on the local covariance $\operatorname{Cov}(R,A\mid B)$ in the retained slice, neither their sign nor the need for FAR is predictable a priori: with residuals non-monotone in $D$, Top-$D$ and Bot-$D$ shift alike (KuaiRand, Table~\ref{tab:replay}), while slices near the random baseline show non-significant $\Delta$ECE under FAR. In all runs, survivor-fit FAR removes the signed gap, and its $90/93\%$ Criteo ECE reduction mirrors the $60\text{--}80\%$ in production, so the recipe carries over across platforms, labels, and selectors.

\section{Conclusion and Operating Principles}
\label{sec:conclusion}
Calibration belongs to a score \emph{and} the population it is applied to, so selection can break even a perfectly calibrated score. In production, ECE rose four- to fivefold between model output and the auction with no model change; one pre-auction FAR map cut it by $60\text{--}71\%$ and improved auction efficiency without significant RPM change, and public replays reproduce the mechanism. We distill five operating principles:
\begin{enumerate}
\item \textbf{Calibrate where scores are used}, not at model output.
\item \textbf{Diagnose every boundary} by splitting signed bias into inherited error, composition shift, and residual shift.
\item \textbf{Match the fix to the error:} upstream refits correct inherited error; only fits on the selected population remove selection bias.
\item \textbf{Refit when the selected population changes}, including through the correction itself.
\item \textbf{Look beyond aggregates:} opposite errors can cancel.
\end{enumerate}
The same failure can arise in recommendation cascades, retrieve-and-rerank, and moderation pipelines; jointly calibrating boundaries and adapting the correction online are natural next steps. The planned release of our adapted datasets and pipeline is intended to provide a reproducible benchmark. Since calibration is usually reported only at model output, offline evaluation understates decision-time error: a score's calibration belongs to the population it serves, not to the model alone.

\bibliographystyle{ACM-Reference-Format}
\begingroup
\raggedright
\bibliography{references}
\endgroup
\appendix

\section{Proofs of the Selection-Induced Calibration Identities}
\label{app:selection-gap-proof}

We derive Equations~\eqref{eq:selection-gap} and~\eqref{eq:decomposition}
step by step and prove the transfer conditions of Section~\ref{sec:gap}.
Neither identity assumes that the input score is calibrated.
Throughout, $Y\in\{0,1\}$,
$S\in[0,1]$, $A\in\{0,1\}$, $R=Y-S$, and $B$ is any partition of $S$'s range,
so $B$ is a function of $S$. We hold this partition fixed before and after
selection. Since $R\in[-1,1]$, all required moments exist. The first identity
isolates residual change within a bin; the second separates its aggregate
effect into inherited error, composition shift, and residual shift.

\subsection{Conditional Residual Identity Under Selection}

\begin{proposition}\label{prop:selection-gap}
For every bin $b$ with $\Pr(B{=}b)>0$ and
$\pi_b=\Pr(A{=}1\mid B{=}b)>0$,
\[
\begin{aligned}
\mathbb{E}[R\mid A{=}1,B{=}b]=\mathbb{E}[R\mid B{=}b]
\\
\quad+\frac{\operatorname{Cov}(R,A\mid B{=}b)}{\pi_b}.
\end{aligned}
\]
\end{proposition}

\begin{proof}
Fix such a bin $b$, and abbreviate its input and survivor mean residuals by
$r_b=\mathbb{E}[R\mid B{=}b]$ and
$r_b^A=\mathbb{E}[R\mid A{=}1,B{=}b]$.

\emph{Step 1 (express selected residual mass).} Multiplication by the
indicator $A$ retains the residual only for survivors: $RA=R$ if $A=1$,
and $RA=0$ otherwise. Taking the conditional expectation gives
\[
\mathbb{E}[RA\mid B{=}b]=\pi_b r_b^A.
\]
This identity also holds when every candidate in the bin survives.

\emph{Step 2 (expand conditional covariance).} By definition,
\[
\begin{aligned}
\operatorname{Cov}(R,A\mid B{=}b)
&=\mathbb{E}[RA\mid B{=}b]\\
&\quad-\mathbb{E}[R\mid B{=}b]\,
         \mathbb{E}[A\mid B{=}b].
\end{aligned}
\]
Because $A$ is binary, $\mathbb{E}[A\mid B{=}b]=\pi_b$. Hence
\[
\operatorname{Cov}(R,A\mid B{=}b)
=\mathbb{E}[RA\mid B{=}b]-r_b\pi_b.
\]

\emph{Step 3 (isolate the survivor residual).} Substituting Step~1 yields
\[
\operatorname{Cov}(R,A\mid B{=}b)
=\pi_b r_b^A-\pi_b r_b
=\pi_b(r_b^A-r_b).
\]
Dividing by $\pi_b>0$ and adding $r_b$ to both sides gives
\[
r_b^A=r_b+\frac{\operatorname{Cov}(R,A\mid B{=}b)}{\pi_b}.
\]

\emph{Step 4 (restore the conditional-expectation notation).} Replacing
$r_b^A$, $r_b$, and $\pi_b$ by their definitions gives
\[
\begin{aligned}
\mathbb{E}[R\mid A{=}1,B{=}b]
&=\mathbb{E}[R\mid B{=}b]\\
&\quad+\frac{\operatorname{Cov}(R,A\mid B{=}b)}
                {\Pr(A{=}1\mid B{=}b)},
\end{aligned}
\]
which is Equation~\eqref{eq:selection-gap}. If $\pi_b=0$, a survivor mean in
that bin is undefined and this conditional identity is not asserted.
\end{proof}

\subsection{Decomposition of Aggregate Survivor Bias}
\label{app:decomposition-proof}

Assume $\Pr(A{=}1)>0$. Sum over bins with $p_b=\Pr(B{=}b)>0$, and let
$p_b^A=\Pr(B{=}b\mid A{=}1)$. For any bin with $\pi_b=0$, we have $p_b^A=0$;
for the sums below, set $r_b^A=r_b$ in such bins as a bookkeeping convention
only. This does not assign a conditional mean to an empty survivor population;
its weighted contribution is zero regardless of the assigned value.

\begin{proposition}\label{prop:decomposition}
The survivor mean residual satisfies
\[
\begin{aligned}
\mathbb{E}[R\mid A{=}1]
&=\sum_b p_b r_b+\sum_b(p_b^A-p_b)r_b\\
&\quad+\sum_b p_b^A(r_b^A-r_b).
\end{aligned}
\]
\end{proposition}

\begin{proof}
\emph{Step 1 (average over survivor bins).} By the law of total expectation
on the survivor population,
\[
\mathbb{E}[R\mid A{=}1]=\sum_b p_b^A r_b^A.
\]
The same law on the input population gives
\[
\mathbb{E}[R]=\sum_b p_b r_b.
\]

\emph{Step 2 (separate within-bin change).} Write each survivor residual as
$r_b^A=r_b+(r_b^A-r_b)$ and distribute the survivor bin weight:
\[
\begin{aligned}
\sum_b p_b^A r_b^A
&=\sum_b p_b^A\big[r_b+(r_b^A-r_b)\big]\\
&=\sum_b p_b^A r_b+\sum_b p_b^A(r_b^A-r_b).
\end{aligned}
\]
The first sum keeps input bin residuals but uses survivor bin shares.
The second measures how selection changes residuals inside each bin.

\emph{Step 3 (separate composition from inherited error).} Add and subtract
the input weighted residual $\sum_b p_b r_b$:
\[
\begin{aligned}
\sum_b p_b^A r_b
&=\sum_b p_b r_b+\sum_b p_b^A r_b-\sum_b p_b r_b\\
&=\sum_b p_b r_b+\sum_b(p_b^A-p_b)r_b.
\end{aligned}
\]
Thus, changing bin shares affects the contribution of an existing residual
$r_b$ through the weight difference $p_b^A-p_b$.

\emph{Step 4 (combine the terms).} Substituting Step~3 into Step~2 and then
into Step~1 yields
\[
\begin{aligned}
\mathbb{E}[R\mid A{=}1]
&=\underbrace{\sum_b p_b r_b}_{\text{inherited error } E}\\
&\quad+\underbrace{\sum_b(p_b^A-p_b)r_b}_{\text{composition shift } C}\\
&\quad+\underbrace{\sum_b p_b^A(r_b^A-r_b)}_{\text{residual shift } W},
\end{aligned}
\]
which is Equation~\eqref{eq:decomposition}.

\emph{Step 5 (identify the covariance contribution).} Applying
Equation~\eqref{eq:selection-gap} in every bin with $\pi_b>0$ gives
\[
\begin{aligned}
W
&=\sum_{b:\pi_b>0}p_b^A
 \frac{\operatorname{Cov}(R,A\mid B{=}b)}{\pi_b}\\
&=\frac{\sum_b p_b\operatorname{Cov}(R,A\mid B{=}b)}
       {\Pr(A{=}1)}.
\end{aligned}
\]
The second equality uses Bayes' rule,
$p_b^A=p_b\pi_b/\Pr(A{=}1)$. Bins with $\pi_b=0$ have zero conditional
covariance because $A=0$ almost surely there, so they contribute zero to
the final sum. Hence the residual shift $W$ is a weighted within-bin
selection covariance, not a change in bin composition.
\end{proof}

\subsection{Calibration Consequences}

If $r_b=0$ for every input bin, then $E=C=0$ in
Equation~\eqref{eq:decomposition}, and Equation~\eqref{eq:selection-gap}
gives $r_b^A=\operatorname{Cov}(R,A\mid B{=}b)/\pi_b$ in every surviving
bin: all remaining bin error is residual shift $W$.
In particular, exact input calibration implies these zero bin residuals
by the tower property:
\[
\mathbb{E}[R\mid B]
=\mathbb{E}\big[\mathbb{E}[R\mid S]\mid B\big]=0.
\]
By contrast, $\mathbb{E}[R]=0$ alone only makes $E$ zero and need not
eliminate $C$. The decomposition concerns signed residuals; taking absolute
bin residuals for ECE does not preserve its additive form.

\subsection{Transfer Conditions}
\label{app:transfer}

Take the finest partition, $B=S$, as in Section~\ref{sec:gap}. If $S$ is
calibrated, $\mathbb{E}[R\mid S]=0$, so by
Proposition~\ref{prop:selection-gap} survivors are calibrated if and only if
$\operatorname{Cov}(R,A\mid S)=\mathbb{E}[RA\mid S]=0$. Let $U$ denote
randomness independent of $(X,Y)$. Each condition of Section~\ref{sec:gap}
implies $\mathbb{E}[RA\mid S]=0$:
\begin{enumerate}
\item \emph{Score-only selection}, $A=f(S,U)$: then $A\perp Y\mid S$, a special
case of condition~3.
\item \emph{Same-information selection}, $S=\mathbb{E}[Y\mid X]$ and $A=f(X,U)$:
since $U\perp Y\mid X$, $\mathbb{E}[R\mid X,A]=\mathbb{E}[Y\mid X]-S=0$, and as
$S$ is a function of $X$, $\mathbb{E}[RA\mid S]
=\mathbb{E}\big[A\,\mathbb{E}[R\mid X,A]\mid S\big]=0$.
\item \emph{No extra outcome information}, $A\perp Y\mid S$: given $S$, $R$ is
a function of $Y$, so $A\perp R\mid S$ and
$\mathbb{E}[RA\mid S]=\mathbb{E}[R\mid S]\Pr(A{=}1\mid S)=0$.
\end{enumerate}
Because $B$ is a function of $S$, the tower property then gives
$\operatorname{Cov}(R,A\mid B)=0$ for any coarser fixed bins as well.

\paragraph{Sign of the gap.} With $S$ calibrated, $\mathbb{E}[R\mid B]=0$,
so the survivor residual has the sign of $\operatorname{Cov}(R,A\mid B)$.
A positive covariance means the selector preferentially retains items whose
outcomes exceed their scores, so $S$ under-predicts on survivors; a negative
covariance produces over-prediction, as at production filtering
(Table~\ref{tab:filtering}).

\begin{example}
Take a bin at $S\approx 0.10$ containing $1{,}000$ candidates with $100$
positives, so $\mathbb{E}[R\mid B]=0$. Suppose a selector retains $200$ of
them ($\pi_b=0.2$) and, using signals correlated with $Y$, those $200$ contain
$40$ positives. Then
$\operatorname{Cov}(R,A\mid B)=\mathbb{E}[RA\mid B]=(40-0.1\cdot 200)/1000=0.02$,
and Proposition~\ref{prop:selection-gap} gives a survivor residual of
$0.02/0.2=0.10$: survivors are positive at rate $0.20$ against a prediction
of $0.10$, although $S$ is calibrated on the full bin.
\end{example}
\end{document}